\documentclass[utf8]{frontiersSCNS} 

\usepackage{url,hyperref,lineno,microtype}
\usepackage[onehalfspacing]{setspace}
\usepackage{blindtext}
\usepackage{enumerate}
\usepackage{amsmath}
\usepackage{graphicx}
\usepackage{float}
\usepackage{textcomp}
\usepackage{footnote}
\usepackage{comment}
\usepackage{color,soul}
\usepackage{amsfonts}
\usepackage{algorithm}
\usepackage{longtable}
\usepackage{arevmath}     
\usepackage[noend]{algpseudocode}
\usepackage{wrapfig}
\usepackage{gensymb}
\usepackage{lipsum}  
\usepackage{makecell}
\usepackage{tabto}
\def\BibTeX{{\rm B\kern-.05em{\sc i\kern-.025em b}\kern-.08em
    T\kern-.1667em\lower.7ex\hbox{E}\kern-.125emX}}
\newcommand{\ts}{\textsuperscript}

\newcommand\tabn[1][1cm]{\hspace*{#1}}

\makeatletter
\def\ALG@special@indent{%
    \ifdim\ALG@thistlm=0pt\relax
        \hskip-\leftmargin
    \else
        \hskip\ALG@thistlm
    \fi
}
\newcommand{\Thread}[1]{\item[]\noindent\ALG@special@indent \textbf{Thread:}\ #1}
\newcommand{\EndThread}{\item[]\noindent\ALG@special@indent \textbf{End thread}}
\newcommand{\Space}{\item[]\noindent\ALG@special@indent}
\makeatother

\def\keyFont{\fontsize{8}{11}\helveticabold }
\def\firstAuthorLast{Machado and McGinnity} 
\def\Authors{Pedro Machado\,$^{1,*}$  and T.M. McGinnity\,$^{1,2}$}
\def\Address{$^{1}$Computational Neuroscience and Cognitive Robotics Group, School of Science and Technology, Nottingham Trent University,Nottingham, NG11 8NS, United Kingdom \\
$^{2}$Intelligent Systems Research Centre, Ulster University, BT48 7JL, N. Ireland}
\def\corrAuthor{Pedro Machado}

\def\corrEmail{pedro.baptistamachado@ntu.ac.uk}

\begin{document}



\onecolumn
\firstpage{1}

\title[Exploiting High Quality Tactile Sensors...]{Exploiting High Quality Tactile Sensors for Simplified Grasping} 

\author[\firstAuthorLast ]{\Authors} 
\address{} 
\correspondance{} 

\extraAuth{}

\maketitle

\begin{abstract}

Robots are expected to grasp a wide range of objects varying in shape, weight or material type. Providing robots with tactile capabilities similar to humans is thus essential for applications involving human-to-robot or robot-to-robot interactions, particularly in those situations where a robot is expected to grasp and manipulate complex objects not previously encountered. A critical aspect for successful object grasp and manipulation is the use of high-quality fingertips equipped with multiple high-performance sensors, distributed appropriately across a specific contact surface.

In this paper we present a detailed analysis of the use of two different types of commercially available robotic fingertips (BioTac and WTS-FT), each of which is equipped with multiple sensors distributed across the fingertips' contact surface. We further demonstrate that, due to the high performance of the fingertips, a complex adaptive grasping algorithm is not required for grasping of everyday objects. We conclude that a simple algorithm based on a proportional controller will suffice for many grasping applications, provided the relevant fingertips exhibit high sensitivity. In a quantified assessment, we also demonstrate that, due in part to the sensor distribution, the BioTac-based fingertip performs better than the WTS-FT device, in enabling lifting of loads up to 850g, and that the simple proportional controller can adapt the grasp even when the object is exposed to significant external vibrational challenges.

\tiny\keyFont{grasping, tactile sensing, WTS-FT fingertips, BioTac fingertips, adaptive grasp}
\end{abstract}



\section{Introduction} \label{chap1:introduction}
Humans use their five senses (\textit{i.e.} touch, sight, hearing, smell and taste) for perceiving the world. The sense of touch is fundamental in enabling humans to perceive the surrounding environment  \cite{Yeon2017}.  In particular, humans use touch for identifying an object's physical properties, which are then used for fine and precise object manipulation \cite{Yeon2017}. Humans have about 241 mechanoreceptors per square centimetre across different skin layers (cutaneous cues) for perceiving fine textural details. Mechanoreceptors in joints and tendons (kinesthetic cues) assist in geometric shape perception \cite{Rouhafzay2019}. 

Unlike humans, current robots have limited capability to perceive the world using the sense of touch and are therefore significantly restricted in performing everyday tasks, such as opening/closing bottles, opening/closing doors, or complex object manipulation. Robots also exhibit difficulties in grasping objects varying in shape, size and weight, particularly if the robot has no prior knowledge of the object or task in question. Because of its importance in sensory perception and action, the scientific community has been actively researching the development of the sense of touch for robots, but it currently lags behind advances in robot vision.

In this paper a comparative analysis of adaptive grasping and object handover using two types of commercially available tactile fingertips (fitted with different types of tactile sensing technologies) is provided. The two tactile fingertips were chosen as they are representative of devices in the moderate (WTS-FT) and high cost (BioTac)  areas of commercially available sensors respectively. It is demonstrated that, due to the high performance of the BioTac fingertips, a complex adaptive grasping algorithm is not required for many grasping situations and a simple proportional controller will suffice, even when the object is subjected to significant perturbations (vigorous. shaking). The two types of fingertips are compared under the same test scenarios, where the object is placed in a target position and the robot arm moves toward the object, grasps it and transfers it to another location. The degree to which each set of fingertips can accommodate different weights and differently shaped loads is also assessed. The proposed adaptive grasping approach is independent of the type of fingertips being used, and exploits control of the three fingers used (\textit{i.e.} thumb, index and ring) for grasping the objects. More details about the adaptive grasp, test scenarios and objects used for the grasp experiments are given in section \ref{chap3:methodology}. 
The paper structure is as follows: background information and related works are presented in section~\ref{chap2:related_work}; the methodology is explained in section~\ref{chap3:methodology}; experimental results are found in  section~\ref{chap4:results}; an analysis of the results  may be seen in section~\ref{chap5:analysis}; section~\ref{chap6:discussion} presents the discussion and conclusion; and future work is discussed in section~\ref{chap7:future}.

\section{Related work} \label{chap2:related_work}
The development of humanoid robots and/or robotic arms capable of sensing and perceiving the environment has attracted the interest of the scientific community for many decades. Recent trends in robotics illustrate the need for robots capable of grasping and manipulating complex shaped or delicate objects in resilient environments (\textit{i.e.} farming, mining, medicine, manufacturing, nuclear and space) \cite{Government2018}. Artificial sensing perception aims to identify physical object properties (\textit{i.e.} deformability, textural characteristics, stiffness, temperature, shape, weight, etc) which then can be used for pose estimation, improving the grasp quality and successful object manipulation. Chi \textit{et al.} discussed state-of-the-art tactile sensing technologies (including capacitive, piezoresistive, piezoelectric, optical and inductive) and these are summarised in
Table~\ref{tab:tranducers} ( \cite{Chi2018}).

\begin{table}[htb!]
\centering
\caption{Transduction mechanisms and its advantages and disadvantages \protect\cite{Chi2018} \\}\label{tab:tranducers}
\begin{tabular}{|c|c|c|}
\hline
\begin{tabular}[c]{@{}c@{}}Transduction\\ Mechanisms\end{tabular} & Advantages & Disadvantages \\ \hline
Capacitive & \begin{tabular}[c]{@{}c@{}}High sensitivity\\ High spatial resolution \\ Large dynamic range\\ Temperature independent\end{tabular} & \begin{tabular}[c]{@{}c@{}}Stray capacitance\\ Complex measurement circuit\\ Cross-talk between elements\\ Susceptible to noise \\ Hysteresis\end{tabular} \\ \hline
Piezoresistive & \begin{tabular}[c]{@{}c@{}}Simple construction\\ High spatial resolution\\ Low cost\\ Suitable for VLSI\end{tabular}                & \begin{tabular}[c]{@{}c@{}}Hysteresis\\ High power consumption\\ Lack of reproducibility\end{tabular} \\ \hline
Piezoelectric & \begin{tabular}[c]{@{}c@{}}High-frequency response\\ High accuracy \\ High sensitivity\\ High dynamic range\end{tabular} & \begin{tabular}[c]{@{}c@{}}Poor spatial resolution\\ Charge leakages\\ Dynamic sensing only\end{tabular}  \\ \hline
Optical & \begin{tabular}[c]{@{}c@{}}Good reliability\\ Wide sensing range\\ High repetability\\ High spatial resolution\end{tabular} & \begin{tabular}[c]{@{}c@{}}Non-conformable\\ Bulky in size\\ Susceptible to temperature\\ or misalignment\end{tabular} \\ \hline
Inductive & \begin{tabular}[c]{@{}c@{}}Linear output\\ High sensitivity\\ High power output\\ High dynamic range\end{tabular} & \begin{tabular}[c]{@{}c@{}}Low-frequency response\\ Poor reliability\\ More power consumption\end{tabular} \\ \hline
\end{tabular}
\end{table}
Kappassov \textit{et al.} \cite{Kappassov2015} reviewed manipulation and grasping applications that involve touch perception and discussed pros and cons of each technique. Kappassov \textit{et al.} focused his report on the analysis of the algorithms including grasp stability estimation, tactile object recognition, tactile servoing and force control. Despite the relevant contribution of Kappassov \textit{et al.} \cite{Kappassov2015}, the focus of the present paper is to compare state-of-the-art commercial-of-the-shelf fingertips using a single common control approach for uniformity, as opposed to comparing control algorithms-sensor combinations. \\
Bhandari and Lee \cite{Bhandari2019} proposed an algorithm to control a robotic hand, acquiring pressure and vision data. Their algorithm uses TakkTile fingertips \cite{Takktile}, which are low cost tactile sensors using micro-electromechanical systems (MEMS) barometers as pressure sensors, capable of detecting forces arising from masses of up to hundreds of grams \cite{Bhandari2019}. Although the TakkTile sensors are low cost, the fingertips are very limited, because they are composed of a single sensor, making it impossible to extract relevant information about the object's spatial or textural properties. Another limitation of the proposed algorithm is that the camera must be perfectly aligned for detecting the object deformation. In contrast, in the present paper, the simple proportional control algorithm exploits multiple sensors per fingertip to enable adaptation of the grasp, while the object is being manipulated. Chorley \textit{et al.} \cite{Chorley2009} proposed a biologically inspired tactile sensor composed of a thin flexible rubber skin, a highly compliant polymer melt blend and a camera facing an array of tracking markers (both the camera and the tracking markers are on the internal structure of the fingertip). The fingertip proposed in \cite{Chorley2009}, is capable of encoding edge information which is crucial for the fine manipulation and exploration of objects. Such capabilities would be desirable when grasping and/or exploring objects; however the sensors are not commercially available.

Kerr \textit{et al.} (\cite{Kerr2013}, \cite{Kerr2014}, \cite{Kerr2018}) proposed the use of the BioTac fingertips from \cite{biotac}; these are robotic fingertips composed of 19 electrodes, AC/DC pressure and temperature transducers. Their research exhibits extensive work on material classification using the BioTac fingertips and different machine learning (ML) and artificial neural networks (ANN) techniques. The results in \cite{Kerr2018} are particularly interesting, because the solution proposed for material classification using the BioTac fingertips outperforms humans by over 16\%, although this is in the context of tactile only (no vision) exploration. Gomez-Eguiluz \textit{et al.} \cite{GomezEguiluz2018} proposed a recursive multimodal (vibration and thermal) tactile material identification approach using the BioTac fingertips and show it is possible to accurately classify the material type in about 0.28 seconds. In \cite{GomezEguiluz2019} the same authors incorporated the use of an effort controller to adapt the grasp forces, using BioTac sensors, when perturbations are detected. The algorithm classifies the force gradient to decide when to release the object in a handover context. Ottenhaus \textit{et al.} (\cite{Ottenhaus2016}, \cite{Ottenhaus2018}) proposed a method for haptic exploration through implicit surface modelling. WTS-FT fingertips \cite{wts-ft} composed of 32 capacitive cells uniformly distributed across the contact surface were used for collecting the tactile data (\cite{Ottenhaus2016}, \cite{Ottenhaus2018}). These authors expanded the work initially presented in \cite{Ottenhaus2016} to propose a contact and surface orientation sensor concept, again for surface reconstruction, with a claimed mean reconstruction accuracy of 3.8mm (\cite{Ottenhaus2018}). Ke \textit{et al.} (\cite{Ke2019}) presented a new anthropomorphic fingertip composed of a force sensing resistor (FSR) and polyvinylidene (PVDF) sensor modules with a claimed force measurement error of less than 5\% , capable of distinguishing 5 different textures with an accuracy above 94\%. More recently, Naganbandi \textit{et al.} \cite{Nagabandi2019} proposed an online planning approach with deep dynamic models for in-hand object manipulation using the BioTac SP sensors.  \cite{Nagabandi2019} reported that the model was able to coordinate multiple free-floating objects after learning from a dataset containing only four hours of synthetic videos performing the target tasks. 

In this paper, the authors report on the development of an adaptive grasping algorithm compatible with different types of fingertips and capable of performing active grasp adaptation while moving the object from a source to a destination, while experiencing shaking perturbations. A number of grasp-related issues are investigated including the influence of the fingertips on the overall performance results and the possibility of completing the complex tasks using different types of sensors. The adaptive grasping algorithm was tested on both the BioTac and the WTS-FT fingertips which are commercially available.

\section{Material and Methods} \label{chap3:methodology}
In this section, relevant aspects of the methodology are reported.
\subsection{Robotic platform} \label{chap3.1:robotic_platform}
\begin{figure} [htb!]
	\begin{center}
	\includegraphics[width=0.8\textwidth]{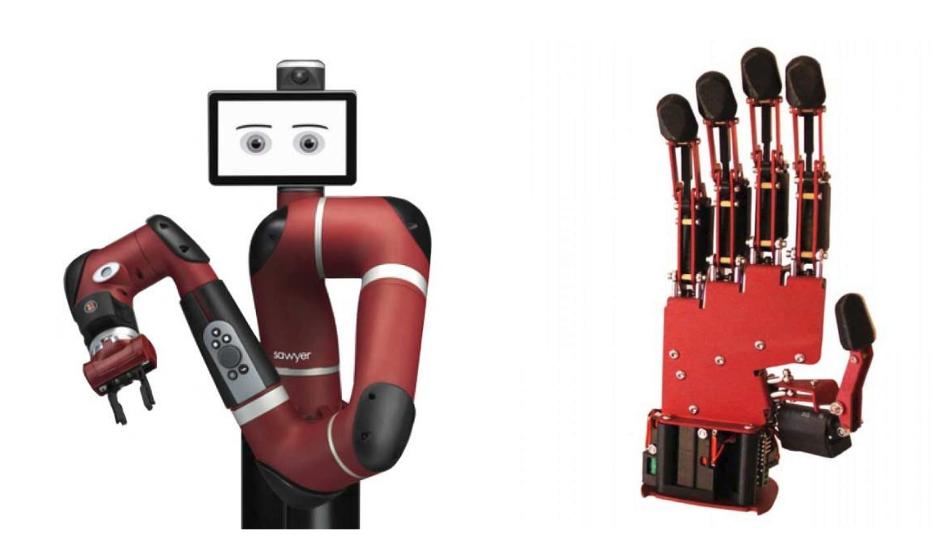}
	\end{center}
	\caption{Sawyer robotic Arm \cite{Sawyer} and AR10 Robotic Hand \cite{ar10}}  \label{fig:sawyer_ar10}
\end{figure}

The Sawyer robot from \cite{Sawyer} is a collaborative robot that was designed for performing complex and accurate object pick-and-place tasks and was used in this work. Its arm (Figure~\ref{fig:sawyer_ar10}) has 7 degrees of freedom, maximum reach of 1.26m, typical speed of 1.5 m/s and can lift payloads with a weight of up to 4Kg.
An AR10 robotic hand from \cite{ar10} was installed on the Sawyer robot and used in this work for grasping objects. The AR10 robotic hand (Figure~\ref{fig:sawyer_ar10}) is a low-cost hand that features 10 degrees of freedom. As the WTS-FT fingertips are relatively large, and there isn't sufficient space for operating all the AR10 fingers when the WTS-FT fingertips are installed on the index and ring fingers, the middle and small fingers were removed to allow for free movement of the remaining fingers.

\subsection{Fingertips utilised} \label{chap3.2:fingertips}
Two types of fingertips were used, namely the BioTac SP and WTS-FT sensors. The specifications of each type of fingertip are discussed below.
\subsubsection{BioTac SP fingertips } \label{chap3.1:biotac_sensors}
\begin{figure} [htb!]
	\includegraphics[width=1.0\textwidth]{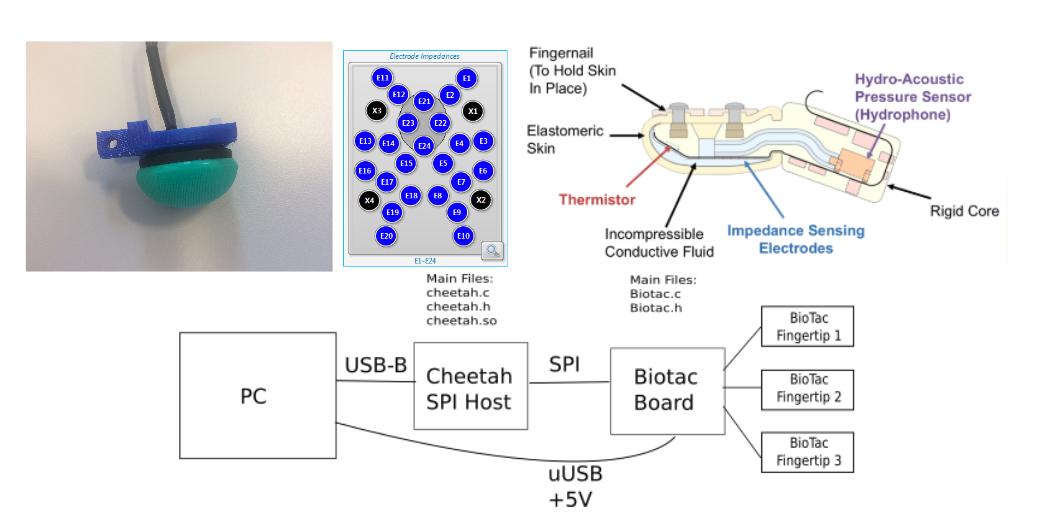}
	\caption{BioTac SP fingertip. Picture of the fingertip (top left), sensors map (top centre),
	Cross section view \protect\cite{biotac} (top right), system setup (bottom)}  \label{fig:biotac}
\end{figure}

The Biotac SP fingertip from \cite{biotac} is a small (26x20x25mm) fingertip composed of an array of tactile sensors with a weight of 9.5g, capable of detecting information similar to human fingers (\textit{i.e.} forces, micro-vibrations and temperature). 
Figure~\ref{fig:biotac} depicts the BioTac SP sensor, and shows the fingertip's sensor distribution.

Three sensory modalities are measured via three distinct types of transducers:
\begin{itemize}
    \item 24 electrodes for detecting the synthetic skin deformation and forces applied directly to the area of contact.
    \item AC/DC pressure (PAC/PDC) transducer for detecting the forces applied directly (area populated with electrodes) and indirectly (not detected by the electrodes) to the synthetic skin of the sensor.
    \item AC/DC temperature (TAC/TDC) thermister for the detection of thermal gradients when in contact with different types of materials.
\end{itemize} 
The sensors are distributed across the curved surface of the BioTac SP fingertips for improving the contact between the synthetic skin and the object. The electrodes are strategically positioned in the middle for detecting direct contacts with the object. Vibrations are detected by the DC pressure (PDC) signal (filtered using a built-in low pass filter) and the AC pressure (PAC) signal (filtered using a built-in band pass filter). The BioTac SP fingertips also provide DC temperature (TDC) and AC temperature (TAC) readings, which can be used for calculating the static and dynamic thermal gradients, and which will vary when the fingertip establishes contact with different types of materials. A cross-sectional view of the BioTac fingertips and distribution of the sensors across the BioTac SP fingertip surface are shown (top) in Fig.\ref{fig:biotac}.

The BioTac SP fingertips setup is depicted in Fig.~\ref{fig:biotac} (bottom) where it can be seen that the fingertips connect to a Sensors PC via an SPI Host and a Biotac interface board. Each BioTaC SP fingertip is sampled at 4.4kHz and generates 864 bits per frame. The sampling rate per transducer is shown in Table~\ref{tab:biotac_sampling_rate}.
\begin{table}[htb!]
\centering
\caption{Sampling rate per transducer of the BioTac SP} \label{tab:biotac_sampling_rate}
\begin{tabular}{|c|c|c|}
\hline
Sensor type    & Number of transducers & Sampling Rate      \\ \hline
Electrodes     & 24                & 73Hz per Electrode \\ \hline
AC Pressure    & 1                 & 2.2kHz             \\ \hline
DC Pressure    & 1                 & 73Hz               \\ \hline
AC Temperature & 1                 & 73Hz               \\ \hline
DC Temperature & 1                 & 73Hz               \\ \hline
\end{tabular}
\end{table} Three BioTac SP fingertips were plugged to the AR10 robotic hand, one in each of the three fingers (\textit{i.e.} thumb, index and ring fingers).

\subsubsection{WTS-FT fingertips} \label{chap3.2:wts-ft_sensors}
\begin{figure} [htb!]
	\includegraphics[width=1.0\textwidth]{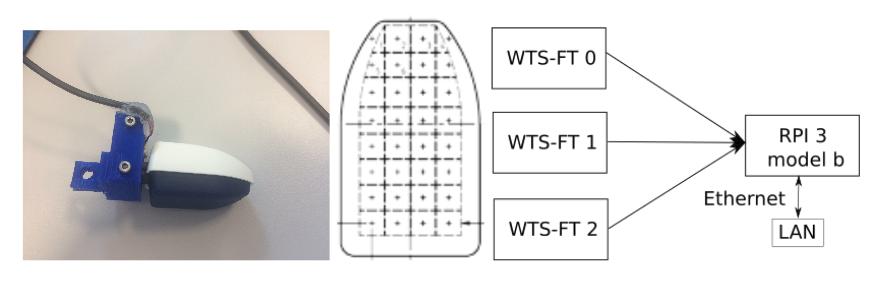}
	\caption{WTS-FT fingertips. fingertip (left), sensors map (centre) and system setup (right)}  \label{fig:wts-ft}
\end{figure}

The WTS-FT fingertip's \cite{wts-ft} tactile sensors measure 52x28x26mm in size, have a weight of 25.8g and are composed of a 4$\times$8 tactile sensing matrix array, exhibiting a human-like sensitivity with measurement speed of up to 400 frames per second. 
Figure~\ref{fig:wts-ft} depicts a WTS-FT sensor (left), sensors map (middle) and the system setup (right).

The WTS-FT fingertip sensors use a USB to serial interface and therefore it was necessary to develop an embedded system for capturing the fingertip data and send them across the Ethernet. A Raspberry Pi 3 model b  was used for capturing the fingertips' data over Ethernet (\cite{RaspberryPiFoundation}). Three WTS-FT fingertips were plugged into the AR10 robotic hand, one in each of the three fingers (\textit{i.e.} thumb, index and ring fingers).

\subsection{Vision system} \label{chap3.4:vision}
The Sawyer robot comes equipped with 2 cameras, one above the display installed on the top of the Sawyer Robot and another on the wrist. Despite having two cameras, the authors concluded that the position and type of the cameras were not ideal. Therefore, the authors installed a state-of-the-art Intel RealSense D435 depth camera from \cite{Intel} and three generic web cameras for capturing different angles of the grasp. The D435 depth camera was installed on the top of the Sawyer robot for capturing the image from above and could also be used for classifying objects using RGB and depth information. 
The four cameras were installed as depicted in Figure~\ref{fig:setup}. Cameras C1, C2 and C3 are the generic cameras that were installed on the top of the table facing the pick-up position, while the D435 depth camera was installed facing the table (see Figure~\ref{fig:setup}).

\subsection{Grasped Objects} \label{chap3.5:grasp_object}
Five objects were select for measuring the quality of the grasp. Each object was selected to assess different abilities when the specific object was being grasped. Each object is manually placed in the grasping region without aligning the object to the robotic hand. The selected objects were:

\begin{itemize}
    \item Plastic cup - a deformable plastic object with the dimensions 9.5$\phi_0\times$ 5.3$\phi_1\times$ 12.5 cm used to evaluate the degree of deformation.
    \item Tea cup - ceramic non-deformable object with the dimensions 8$\times$8$\times$8 cm.
    \item Can - cylindrical object with the dimensions 7.9$\times$7.9$\times$25.9 cm made out of plastic and thin foil used to evaluate the grasp efficiency.
    \item Water bottle - irregularly shaped, non-solid plastic object with the dimensions 7.3$\times$8.7$\times$10 cm used to evaluate the grasp efficiency.
    \item Cubic box - this is a wooden cubic object  with non-parallel sides with the dimensions 8$\times$8$\times$8 cm, chosen as a challenging object to evaluate the contact of the three fingers when an object with flat but non-parallel sides was grasped
\end{itemize}
The weight of each object was adjusted by adding either water or sugar as appropriate. A fail was recorded where the object is dropped at least 50\% of the times (\textit{i.e.} 5 times in 10 repetitions). 

\subsection{Software environment} \label{chap3.5:software}
\begin{figure} [htb!]
	\includegraphics[width=0.6\textwidth]{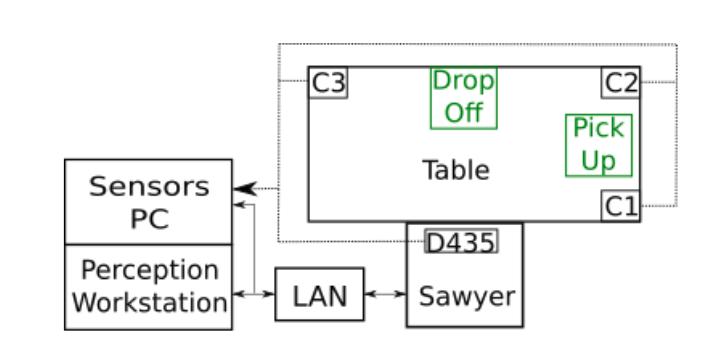}
	\caption{Experimental setup}  \label{fig:setup}
\end{figure}

A distributed computing system composed of a Perception Workstation (a high performance workstation equipped with 20 hyper-threaded cores (Intel Xeon processor), 64 GB of DDR4 and one NVIDIA GPU) was utilised for running the adaptive grasping and vision algorithms. The configuration also included a Sensors PC (a generic PC equipped with an Intel i5 quad-core and 8GB of DDR4) for collecting the sensors' data (\textit{i.e.} fingertips and cameras) and the Sawyer robot PC.
All three computing platforms were running Ubuntu 16.04 LTS \cite{Ubuntu} and the Robotic Operating System (ROS) Kinetic \cite{ROS}. OpenCV 4.0.1 \cite{OpenCV}, a computer vision and machine learning framework was installed on the high performance workstation for processing the images being collected by the Sensors PC. The liberealsense SDK \cite{Realsense} was installed on the Sensors PC for aligning the Red, Green and Blue (RGB) and depth images captured by the D435 depth camera. The Intera SDK \cite{RethinkRobotics} , which is required for communicating with the Sawyer Robot, was installed on both the Perception Workstation and Sensors PC (see Figure~\ref{fig:setup}).

\subsection{Software Developed} \label{chap3.6:software}
Relevant aspects of the  software developed are discussed in this section.
\subsubsection{BioTac SP ROS stack} \label{chap3.6.1:biotac}
The BioTac SP fingertips are state-of-the-art fingertips but unfortunately did not have the required drivers or ROS package. The authors developed appropriate drivers and a ROS stack.

Each sensor of the BioTac SP fingertip's has a resolution of 12 bits and a frequency response of 1.04 kHz, meaning that sensors are sensitive to high frequency noise. A simple low pass filter, described by equation~\ref{eq:low_pass}, was implemented for filtering the BioTac SP fingertips readings. 
\begin{equation} \label{eq:low_pass}
A_n(\Delta_t)=P\times\sum_{i=0}^n A_{n_i}(\Delta t -1)+ (1-P)\times\sum_{i=0}^n r_{n_i}(\Delta t)
\end{equation}
where $A_n(\Delta t)$ is the reading matrix of the $n^{th}$ fingertip for the time step $(\Delta t)$, P is the percentage of the previous readings that should be retained ($0.0\ge P \le1.0$), $A_{n_i}(\Delta t -1)$ is the $i^{th}$ sensor reading value of the the $n^{th}$ fingertip for the time step $(\Delta t -1)$ and $r_{n_i}(\Delta t)$ is the $i^{th}$ sensor reading of the $n^{th}$ fingertip for the time step $(\Delta t)$. 
The modulo of the difference between the A and r matrices (see equation~\ref{eq:modulo}) is computed for normalising the values (sensors with minimal variation will tend to 0 while sensors with a substantial variation will suffer a variation up to 200).
\begin{equation} \label{eq:modulo}
\bar{A_n}(\Delta_t)=\|A_{n}(\Delta t)-r_{n}(\Delta t)\|
\end{equation}
where $\bar{A_n}(\Delta_t)$ is the modulo of the difference between $A_{n}(\Delta t)$ and $r_{n}(\Delta t)$ is the sum of the sensor reading of the $n^{th}$ fingertip. 

The BioTac SP ROS stack is a distributed system where the fingertips data capture functionality runs on the Sensors PC and the data visualisation runs on the Perception Workstation.

\subsubsection{WTS-FT ROS stack} \label{chap3.6.2:wts-ft}
The WTS-FT fingertips had drivers only available for Microsoft Windows. The authors wrote the Linux drivers, created a new WTS-FT ROS stack used for collecting the data generated by WTS-FT fingertips, publishing it into the respective ROS topics and displaying the results on the screen (see Fig.~\ref{fig:wts_cube} and \ref{fig:wts_bottle}). These fingertips have a resolution of 12 bits and a readout rate of 400 frames per second. Unlike the BioTac SP fingertips, no extra filtering was required because the sensor cell's output is linearly proportional to the applied force. The WTS-FT ROS stack is a distributed system, where the fingertips data capture runs on the Raspberry pi 3 model B (refer to Fig.~\ref{fig:wts-ft}) and the data visualisation runs on the Perception PC. 

\subsubsection{Object detection ROS stack} \label{chap3.6.3:object-detection}
The Object Detection ROS stack is a generic package that was written for managing and processing the data collected by the cameras (refer to Fig~\ref{fig:setup}). It is composed of (i) publishers that collect the data from the cameras and publish them on the target topics; (ii) a classification algorithm that classifies the object accordingly to its shape, (iii) visualisation windows that display the images captured from the cameras and classification results and (iv) a recording functionality used for recording the image sequences from the selected topics. The object detection ROS stack is a distributed system that runs on both the Sensors PC (functionality (i)) and on the Perception Workstation (functionalities (ii) to (iv)). 

\subsubsection{Sawyer Grasping Project ROS stack}  \label{chap3.6.4:sawyer-grasping-project}
The Sawyer Grasping Project (SGP) ROS stack (\cite{machado_2019_sawyer}) contains the simple adaptive grasping algorithm. The SGP ROS stack delivers the desirable fingertips type abstraction so that both types of fingertip are tested under the same testing conditions during the object grasping and relocation tasks. This stack includes (i) the main grasping and repositioning Algorithm \ref{alg:main}, (ii) the adaptive grasping  Algorithm \ref{alg:adaptive_grasping}, both shown overleaf, and (iii) the AR10 controller used for controlling the AR10 robotic hand fingers.  

The process of grasping and repositioning an object (Algorithm~\ref{alg:main}) starts when a request for grasping a new object  (sphere, cylinder, cuboid, prism and precision grasps) is received. The robotic arm moves from the home position to the pick up position using one of four pre-loaded grasps (sphere 3 fingers, large diameter,  precision disk and precision sphere) from a pre-grasp library adapted from the grasp database reported by Saubayev \textit{et al.} \cite{Saudabayev2018}. 

A proportional (P) controller, described by equation~\ref{eq:P_controller}, was implemented for controlling the closure speed of the AR10 robotic hand. The P controller takes as input the robotic hand's joints' actual position and returns the next position as described by equation \ref{eq:P_controller}. 

\begin{equation} \label{eq:P_controller}
P_{out}=K_p \times e(t) + p0
\end{equation}

where $P_{out}$ is the output of the proportional controller, $K_p$ is the proportional gain, $p0$ is the controller output with zero error and $e(t)$ is the controller error which is given by equation~\ref{eq:P_controller1},
\begin{equation} \label{eq:P_controller1}
e(t)=SP-PV
\end{equation}

where $SP$ is the set point and $PV$ is the measured process variable. 
The fingertips type abstraction is given through a functionality designated as contact detection. The contact detection takes the readings from the fingertips being used and compares the fingertip's sensor reading to a threshold that was experimentally obtained for each type of fingertip.
Contact is detected (equation~\ref{eq:sensor_value}) when at least 5 sensors $\psi$ (electrode, pressure, temperature in the case of the BioTac SP fingertips or force sensor cells in the case of WTS-FT) of a fingertip exceed a contact threshold $\zeta$. $\zeta$ for the WTS-FT and BioTac was experimentally obtained based on the criteria of (i) lowest object deformation, (ii) highest grasp efficiency rate and (iii) highest slip resistance for a constant finger tracking force of 30N.
The contact detection vector $C$ is a Boolean vector of dimension 3 (equal to the number of fingertips). The contact detection is given by equations~\ref{eq:sensor_value} and \ref{eq:contact_detector}.

\begin{equation} \label{eq:sensor_value}
\Theta^j=
\begin{cases}
 & 1 \textbf{ if } \theta\geq \zeta \\ 
 & 0 \textbf{ if } \theta < \zeta
\end{cases}
\end{equation}

where $\Theta^j$ is the sensor value for fingertip $j$ and $\zeta$ is the experimental threshold. It is considered that sensor $j$ is in contact with the object when $\Theta^j=1$.

\begin{equation} \label{eq:contact_detector}
C^k=
\begin{cases}
 & 1 \textbf{ if } \sum_{j=0}^{z}\ \Theta^j \geq \psi \\ 
 & 0 \textbf{ if } \sum_{j=0}^{z}\ \Theta^j < \psi
\end{cases}
\end{equation}
where $C^k$ is the contact detector value for finger $k$ and $\psi$ the number of activated sensors. It is considered that fingertip $k$ is in contact with the object when $C^k=1$.

The joint target position ($S_{out}$) is a vector containing the target position for each joint and has a dimension of 6 (two joints per finger and a total of 3 fingers).
The target position $S_{out}^i$ for the joint $i$ is given by the following pseudo-code:\\

\textbf{if} ($S_{min}^i\leq P_{out}^i\leq S_{max}^i$) \textbf{and}\ $C^k==0$ \textbf{then}\\
\\
\tabn[1cm] $S_{out}^i=P_{out}^i$\\
\\
\tabn[0.1cm] \textbf{else if} ($P_{out}^i<S_{min}^i$ )\textbf{ and}\ $C^k==0$ \textbf{then}\\
\\
\tabn[1cm] $S_{out}^i=S_{min}^i$\\
\\
\tabn[0.1cm] \textbf{else if} ($ P_{out}^i> S_{max}^i$) \textbf{ and}\ $C^k==0$ \textbf{then}\\
\\
\tabn[1cm] $S_{out}^i=S_{max}^i$\\
\\
\tabn[0.1cm] \textbf{else}\\
\\
\tabn[1cm] $S_{out}^i=S_a^i$\\
\\
\tabn[0.1cm] \textbf{end if}\\


where $S_{min}^i$, $S_{max}^i$, $S_{out}^i$ and $S_a^i$ are the minimum, maximum, target position, actual position for a given joint $i$, respectively.
The target position ($S_{out}^i$) is only updated when a given joint is in between the maximum ($S_{max}^i$) and minimum ($S_{max}^i$) allowed positions and no contact ($C^k=0$) is detected.\\

The grasp is completed when pseudo-code is satisfied for all the joints (\textit{i.e.} $\sum_{i=0}^{joints}S_{out}^i = (S_{max}^i \lor S_{min}^i\lor S_a^i)$). \\

The continuous verification of the contact detection vector and the continuous adaptation of the finger's joints for satisfying the condition $\sum_{i=0}^{joints}S_{out}^i = (S_{max}^i \lor S_{min}^i\lor S_a^i)$ using the the proportional controller, described by equation~\ref{eq:P_controller} is designated as the grasp adaptation. \\

Once an object is grasped, the grasp quality is subsequently assessed by picking the object up, shaking it and transporting it to a drop off position, while continuously performing the grasp adaptation.

\begin{algorithm}[]
\scriptsize
\caption{Main grasping assessment algorithm} \label{alg:main}
\begin{algorithmic}[1]

\Procedure{grasp}{$objects$} 
    \While{True}
        \State $1:\ Idle$ \Comment{1) Idle}
        \State $Select\ object\ to\ grasp$
        \If{$grasp\ object\ request = True$}
            \State $load\ pre-grasp\ offsets;$
            \State $move\ arm\ to\ pick\ up\ point;$
            \State $start\ adaptive\ grasp;$
            \State $load\ Kp1\ values;$
            \State $contact\_vector=get\_contact\_vector;$
            \State $min\_positions=get\_min\_positions;$
            \State $min\_detector=(False,False,False);$
            \While{$3\ fingertips \not= (contact\ \textbf{or} \ min\_position)$}
                \State $contact\_vector=get\_contact\_vector;$
                \State $actual\_position=get\_actual\_position;$
                \For{$finger\ in\ fingers$}
                    \If{$contact\_vector(finger)==False\ \textbf{or}\ min\_detector(finger)\not=min\_position$}
                        \State $target\_position, min\_detector=adaptive\_grasp(actual\_position,Kp1)$
                        \State $close\_hand(target\_position);$
                    \EndIf
                \EndFor
            \EndWhile
            \State $load\ Kp2\ values$
            \State $phase = 0$
            \State $lock = True$ \Comment{2) pick up}
            \While{$phase\not=4$} \Comment{3) Grasp}
                \State $contact\_vector=get\_contact\_vector;$
                \State $actual\_position=get\_actual\_position;$
                \If{$finger\not=contact\ or\ finger\not=min\_positions$}
                        \State $target\_position, min\_detector=Adaptive\_grasp(actual\_position,Kp2);$
                    \EndIf
                \If {$phase==False\ \&\ lock==False$} \Comment{4) lift}
                    \State $task=lift\ object;$
                    \State $lock = True;$
                \ElsIf{$phase==True\ \&\ lock==False$} \Comment{5) shake}
                    \State $task=Shake\ object;$
                    \State $lock = True;$
                \ElsIf{$phase==2\ \&\ lock==False$} \Comment{6) drop-off}
                    \State $task=Move\ arm\ to\ drop\ off\ position;$
                    \State $lock = True;$
                \ElsIf{$phase==3\ \&\ lock==False$} \Comment{7) Release}
                    \State $task=Open\ hand;$
                    \State $lock = 1;$
                \EndIf
                \If {$lock==True$}
                    \State $do\ task\ using\ the\ parameters\ (phase,target\_position);$
                    \If{$task\ status == True$}
                        \State $lock=False;$
                        \State $phase = phase + 1;$
                    \EndIf
                \EndIf
            \EndWhile
        \EndIf
    \EndWhile
\EndProcedure

\end{algorithmic}
\end{algorithm}

The main algorithm, described in Algorithm 1, is composed of the following seven phases: \begin{enumerate}
    \item  The robot is \textbf{idle} awaiting for a new request;
    \item  A new valid request is received and the robot moves the arm to the \textbf{pick up} position;
    \item  The robot \textbf{grasps} the object;
    \item  The object is \textbf{lifted} by the robot;
    \item  The robot \textbf{shakes} the object;
    \item  The object is transported to the \textbf{drop-off} position;
    \item  The robot \textbf{releases} the object and goes to phase \textbf{1)};
\end{enumerate}
The activity encapsulated in phases 4 to 6 constitute the evaluation of the adaptive grasp. The adaptive grasp is a thread that checks the contact detection vector and adjusts (as far as possible) the fingers that may no longer be in contact with the object as a consequence of the type of arm movement, vibrations or object slippage. The main threads of the grasping algorithm are described in Algorithm~\ref{alg:adaptive_grasping}.

\begin{algorithm}[]
\scriptsize
\caption{Adaptive grasping} \label{alg:adaptive_grasping}
\begin{algorithmic}[1]

\Procedure{Grasp}{$object$}
    \State Initialisation
    \Thread 1: Contact detector
        \While{True}
            \State $Collect\ haptic\ fingertips\ raw\ data$
            \State $Display\ fingertips\ results$
            \State $Contact\ detection$
            \State $Publish\ contact\ detection\ vector$
        \EndWhile
    \EndThread
    \Space
     \Thread 2: Haptic fingertips
         \While{True}
            \State $Read\ values\ from\ fingertips$
            \State $Publish\ fingertips\ raw\ data$
        \EndWhile
    \EndThread
    \Space
    \Thread 3: Haptic callback
         \While{True}
            \State $Subscribe\ contact\ detection\ vector$
            \State $Update\ the\ global\ contact\ vector$
        \EndWhile
    \EndThread
    \Space
    \Thread 4: Adaptive Grasp
        \While{True}
            \State $Select\ object\ to\ grasp$
            \If{$close\ hand\ request= True$}
                \State $get\ Kp$
                \State $get\ actual\_position$
                \State $min\_detector=(False, False, False)$
                \For{$finger\ in\ fingers$}
                    \If{$actual\_position(finger)-Kp \ge min\_position$}
                        \State $actual\_position(finger)=actual\_position(finger)-Kp$
                    \Else
                        \State $actual\_position(finger)=min\_position$
                        \State $min\_detector(Finger)=True$
                    \EndIf
                \EndFor
                \State $return (actual\_position, min\_detector)$
            \EndIf
        \EndWhile
    \EndThread
\EndProcedure

\end{algorithmic}
\end{algorithm}

The adaptive grasp algorithm is deliberately kept simple, as the hypothesis was that a simple algorithm would suffice provided high-performance sensor data were utilised. This proved to be the case (see the Results section). The algorithm was however optimised for parallelising all the tasks that could possibly be parallelised (\textit{e.g.} reading fingertips, moving the arm and closing the hand).  

\subsection{Test scenario} \label{chap3.7:test_scenario}
It is important to evaluate both the grasp and the grasp adaptation under realistic conditions.The quality of the grasp was measured using generic tests designed for evaluating the motion and effort described by the Sawyer robot and AR10 hand fitted with either the Biotac SP or WTS-F fingertips (T). The tests also included a range of challenging grasp objects varying in shape, material, weight, texture and size. The test scenarios were divided into two types, namely, performance tests (measuring the motion and effort described by the hand with each type of object) and perturbation tests (evaluating the performance when grasping different types of objects under shaking perturbations). The perturbation test is composed of an acceleration of the speed from 0.1 rad/sec to 0.13 rad/sec, followed by two sequential oscillations in the horizontal axis followed by two sequential oscillations in the vertical axis, each with an amplitude of 10 degrees and a decrease of speed  from the maximum of 0.13rad/sec to the initial 0.1 rad/sec.

Two tests were designed utilised to evaluate the grasps in this work, namely \textbf{slip resistance and touch sensitivity} in this work. 

The \textbf{slip resistance} was measured using a millimetre scale installed on a cylindrical object (\textit{e.g.} water bottle) attached to a spring balance when the object is pulled until it reaches the force threshold. The force threshold was set at a level large enough to cause the object to slip.

The \textbf{touch sensitivity} was measured by pushing a sponge against the sensor under test until contact was detected (i.e. until a value grater than 0 is registered).

\section{Experimental results} \label{chap4:results}
In this section the experimental results obtained when performing the basic grasp performance and performance under object perturbation tests are reported. 

\subsection{Performance tests} \label{chap4.1.1:perfomance_tests}

\begin{figure} [htb!]
	\begin{center}
 \includegraphics[width=0.6\textwidth]{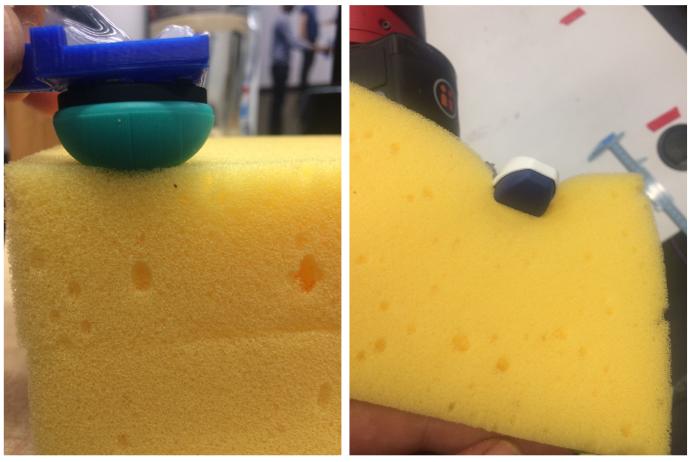}
\caption{Biotac (left) and WTS-FT fingertips touching the sponge. The average of the smallest deformation at which contact was registered was 5mm for the Biotac and 30mm for the WTS-FT Fingertips.}  \label{fig:touch_sensivity}
\end{center}
\end{figure}

The simple adaptive grasping algorithm shown in Algorithm 2 was utilised in all tests of touch sensitivity and slippage analysis. A sponge was used for measuring the minimum deformation required to establish contact with one finger (see Figure~\ref{fig:touch_sensivity}). 

\begin{figure} [htb!]
	\begin{center}
	\includegraphics[width=0.8\textwidth]{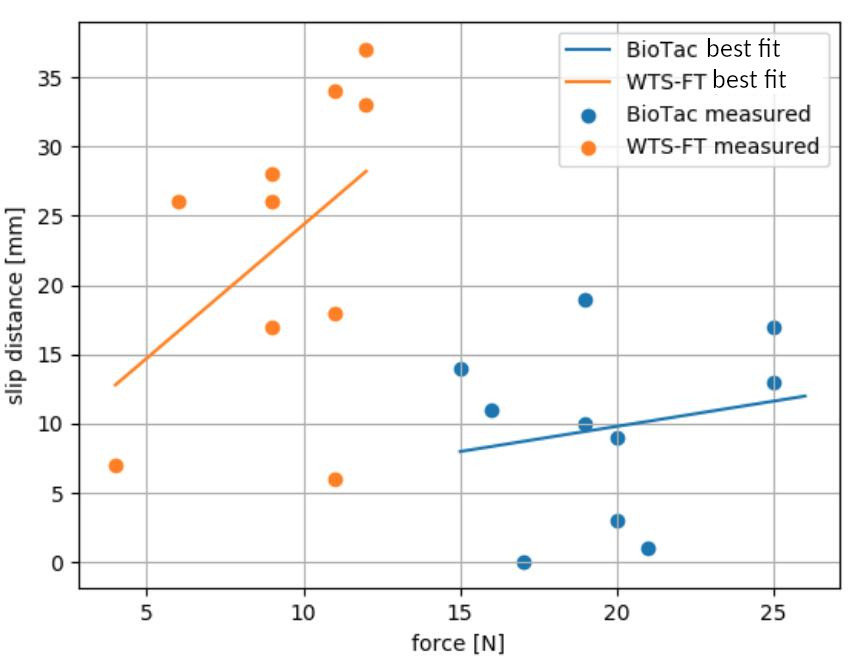}
	\caption{Scatter plot of slippage vs force. The curve fit lines were obtained using the least squares 2\ts{nd} degree polynomial fit.}  \label{fig:force_slip}
	\end{center}
\end{figure}

Figure~\ref{fig:force_slip} presents a scatter plot of slip distance versus force for both the Biotac and WTS equipped hands for the water bottle object. Slippage begins at an earlier force for the WTS sensors, as compared to the Biotac and increases more rapidly as the force is increased.

From the results shown in Figures~\ref{fig:touch_sensivity} and \ref{fig:force_slip} it is clear that the BioTac SP fingertips perform better than the WTS-FT fingertips in terms of touch sensitivity and slip prevention.  The high sensitivity exhibited by the BioTaC devices enables the simple adaptive grasp algorithm to achieve better performance. 
Another factor that contributed to the better performance demonstrated by the BioTac SP fingertips was the circular fingertip shape that improves the contact when touching the object. Unlike the BioTac fingertips that are covered with a skin-like material, the WTS-FT are covered by a rubber-like material that does not adhere very well to the object surface, resulting in weaker grasps and an inability to lift heavier objects. 
\subsection{Performance under object perturbation} \label{chap4.1.2:perturbation_test}
The results of the object perturbation tests are shown in Table~\ref{tab:perturbation_tests}. In each case, the object initially was grasped while empty and subsequently sugar or water (only in the case of the bottle) was added incrementally by 10g/10ml into the object to increase the weight until the object either slipped or dropped. For each weight, the object was then grasped, lifted and shaken (as shown in the yaw and pitch diagrams \textit{e.g.} Figure 13). The maximum weight which the hand/fingertips combination was able to lift and retain, without slippage, while being actively shaken is given in columns 2 and 3 of Table~\ref{tab:perturbation_tests}, for the BioTac and WTS sensors respectively.
The test results show that the BioTac SP fingertips performed equal to or better than the WTS-FT fingertips for all objects and the BioTac SP fingertips enabled the AR10 hand to grasp and lift heavier objects. The main reasons are related to the circular configuration of the BioTac SP fingertips that improves the contact of the fingertips' sensors with the surface of the object. Also the rogusity of the BioTac's skin increases the slip resistance. Note that in the case of the WTS sensors grasping the plastic bottle, once the weight exceeded 500g the sensors could not actually detect contact. This is due to the alteration of the bottle shape and the WTS-FT fingertip configuration.

\begin{table}[htb!]
\centering
\caption{Perturbation tests\\}
\label{tab:perturbation_tests}
\begin{tabular}{|c|c|c|c|}
\hline
 \thead{Description} & \thead{BioTac SP results}& \thead{WTS-FT results} \\
 \hline \hline
 \ Can & 850g  & 300g \\ \hline
 \ Cube Box  & 450g & 350g \\ \hline 
 \ Tea cup  & 500g  & 400g  \\ \hline
 \ Plastic bottle  & 510g & 510g$^*$\\ \hline
 \ Plastic cup   & 350g  & 350g \\ \hline
\end{tabular}
\newline
\\$^*$contact with bottle lost when weight was increased from 500 to 510g due to the shape of the WTS-FT fingertips (see Figure~\ref{fig:wts_bottle})
\end{table}

\subsection{Detailed analysis of fingertip performance}
\begin{figure} [htb!]
	\begin{center}
	\includegraphics[width=1.0\textwidth]{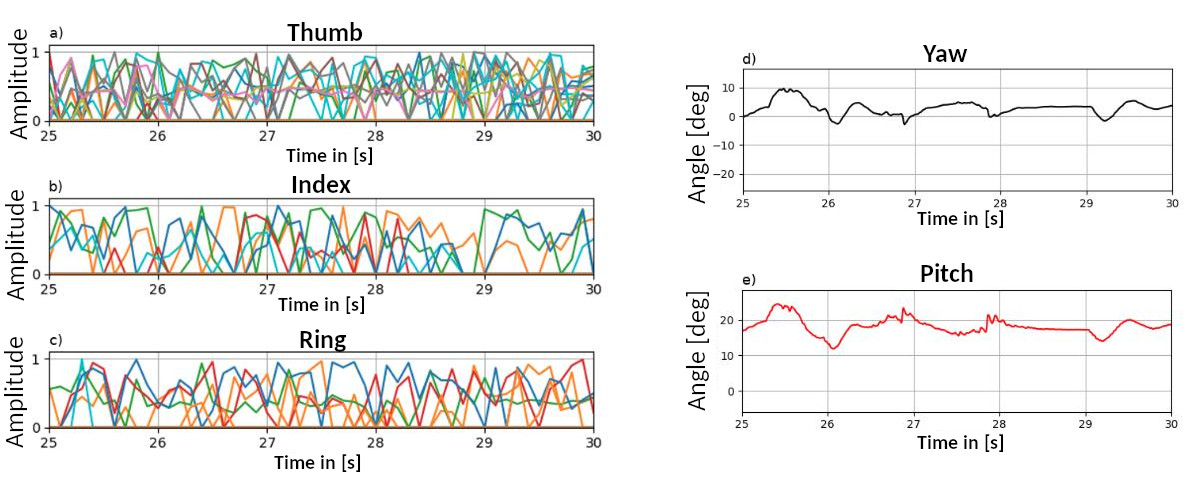}
	\caption{WTS-FT fingertips when grasping the empty cube during the shaking phase. a) shows the amplitudes of the thumb fingertip, b) shows the amplitudes of the index fingertip, c) shows the amplitudes of the ring fingertip, d) the recorded vibrations in yaw while shaking e) the recorded vibrations in pitch while shaking. Video showing the perturbation of the object available at \protect\url{https://zenodo.org/record/3547596} \cite{machado_2019_wts-ft_cube}}  \label{fig:wts_cube}
	\end{center} 
\end{figure}

In this section, examples of  the fingertips’ sensors and vibrations recordings, while utilising the simple adaptive grasping algorithm, are presented and analysed in more detail. The spatial representation is performed using a hot colour map (\textit{i.e.} black [no contact], brown,  red, orange, yellow and white [maximum contact]). An Inertial Measurements Unit (IMU) was installed on the back of the AR10 robotic hand for recording the yaw (horizontal vibrations) and pitch (vertical vibrations).
Figure~\ref{fig:wts_cube} depicts the recordings when grasping the empty cube using the WTS-FT sensors. Each colour refers to a specific sensor (one of 32) for the particular thumb, index or ring finger. 

The plot of the yaw (d) and pitch (e) show the recorded vibrations during the shaking period. The Thumb, Index and Ring Fingertips' sensors activation are shown in a), b) and c), respectively. Contact was detected when at least three sensors were activated (\textit{i.e.} three coloured lines go above 30\%). A full video of the grasp is available in \cite{machado_2019_wts-ft_cube}.

\begin{figure} [htb!]
	\begin{center}
	\includegraphics[width=1.0\textwidth]{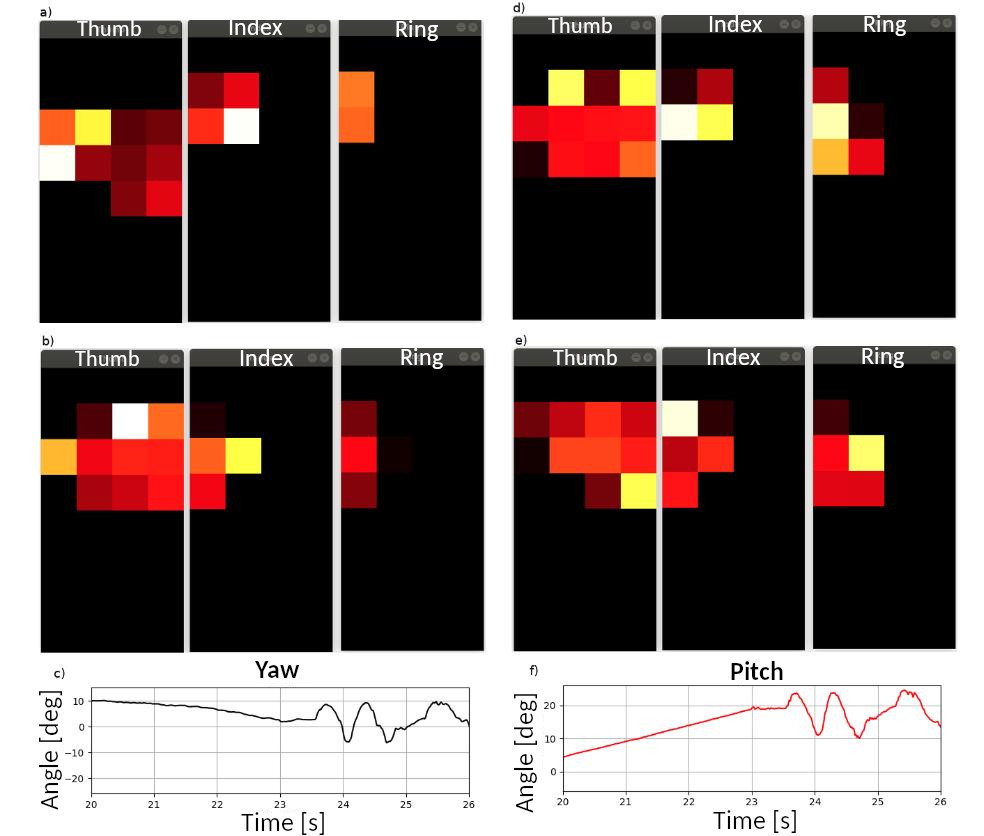}
	\caption{Sequence of the WTS-FT fingertips spacial representation and the recorded vibrations during shaking phase. a) spatial representation at timestamp 21s, b) spatial representation at timestamp 22s, c) plot of the recorded horizontal vibrations (yaw), d) spatial representation at timestamp 23s, e) spatial representation at timestamp 24s and f) plot of the recorded horizontal vibrations (pitch).}  \label{fig:wtf_cube_seq}
	\end{center}
\end{figure}

A spatio-temporal representation of the three fingertips during the shaking phase is shown in Figure~\ref{fig:wtf_cube_seq}.
 The spatial representation of the three fingertip's (Thumb, Index and Ring) sensors were recorded at the timestamps 21, 22, 23 and 24 seconds (a, b, d and e) during the shaking phase. The 4 sequences of the fingertips sensors spacial representation shows grasp adaptation while the object is being shaken. Note that for example only two sensors are detecting pressure in the Ring fingertip at 21s (a), three sensors at 22s (b), five sensors at 23s (d) and five sensors at 24s (e). Furthermore note the variation  in pressure applied to each fingertip's sensor (colour variation). The recorded horizontal (c) and vertical (f) vibrations show the type of movements described by the robotic arm during the shaking phase.
 
\begin{figure} [htb!]
	\begin{center}
	\includegraphics[width=1.0\textwidth]{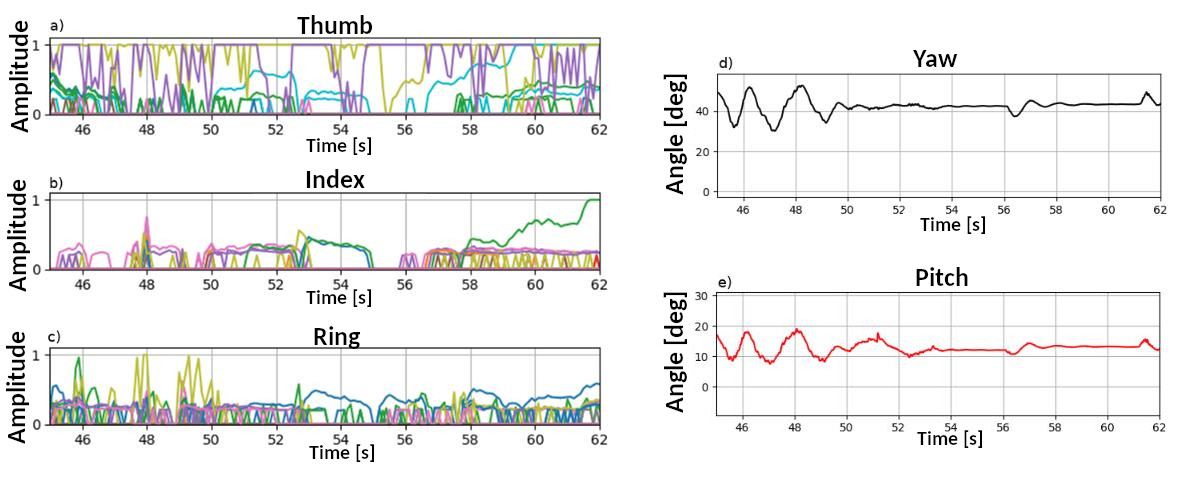}
	\caption{BioTac fingertips when grasping the empty cube during the shaking phase. a) amplitudes of the Thumb fingertip, b) amplitudes of the Index fingertip, c) amplitudes of the Ring fingertip, d) plot of the horizontal vibrations (yaw), e) plot of the horizontal vibrations (pitch). Video showing the perturbation of the object available at \protect\url{https://zenodo.org/record/3547726} \cite{machado_2019_biotac_cube}}  \label{fig:biotac_cube}
	\end{center}
\end{figure}

 Figure~\ref{fig:biotac_cube} depicts equivalent recordings when grasping the empty cube using the BioTac SP fingertips, showing the object vibration during the shaking phase, while the grasp was being adjusted.. 

 The plot of the yaw (d) and pitch (e) show the recorded vibrations during the shaking period. The Thumb, Index and Ring fingertips' sensors activation are shown in a), b) and c), respectively. Contact was detected when when at least five sensors were activated (\textit{i.e.} five coloured lines go above 30\%). 
A full video of the grasp is available in \cite{machado_2019_biotac_cube}.

\begin{figure} [htb!]
	\begin{center}
	\includegraphics[width=1.0\textwidth]{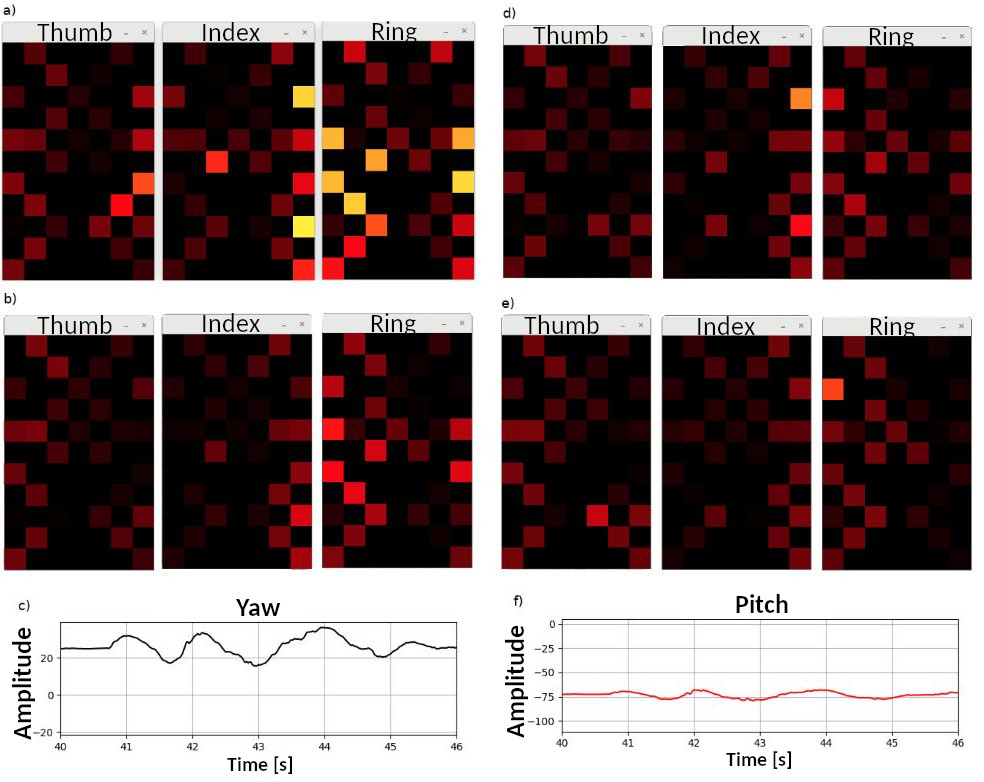}
	\caption{Sequence of the three BioTac SP fingertips spacial representation and the recorded vibrations during shaking phase. a) spatial representation at timestamp 41s, b) spatial representation at timestamp 42s, c) plot of the horizontal (yaw) vibrations, d) spatial representation at timestamp 43s, e) spatial representation at timestamp 44s and f) plot of the vertical (pitch) vibrations.}  \label{fig:biotac_cube_seq}
	\end{center}
\end{figure}

 Figure~\ref{fig:biotac_cube_seq}
highlights the spatial representation of the fingertip's sensors during the shaking phase. The spatial representation of the three fingertip's (Thumb, Index and Ring) sensors were recorded at timestamps 41, 42, 43 and 44 seconds (a, b, d and e) during the shaking phase. The 4 sequences of the fingertips sensors spacial representation's show grasp adaptation while the object is being shook. The recorded horizontal (c) and vertical (f) vibrations show the type of movements described by the robotic arm during the shaking phase. It is also clear that the contact (number of activated sensors per fingertip) between the BioTac SP fingertips (Figure~\ref{fig:biotac_cube_seq}) and the object is much better than the contact between the WTS-FT fingertips (Figure~\ref{fig:wtf_cube_seq}) and the object. 

\begin{figure}
	\begin{center}
	\includegraphics[width=1.0\textwidth]{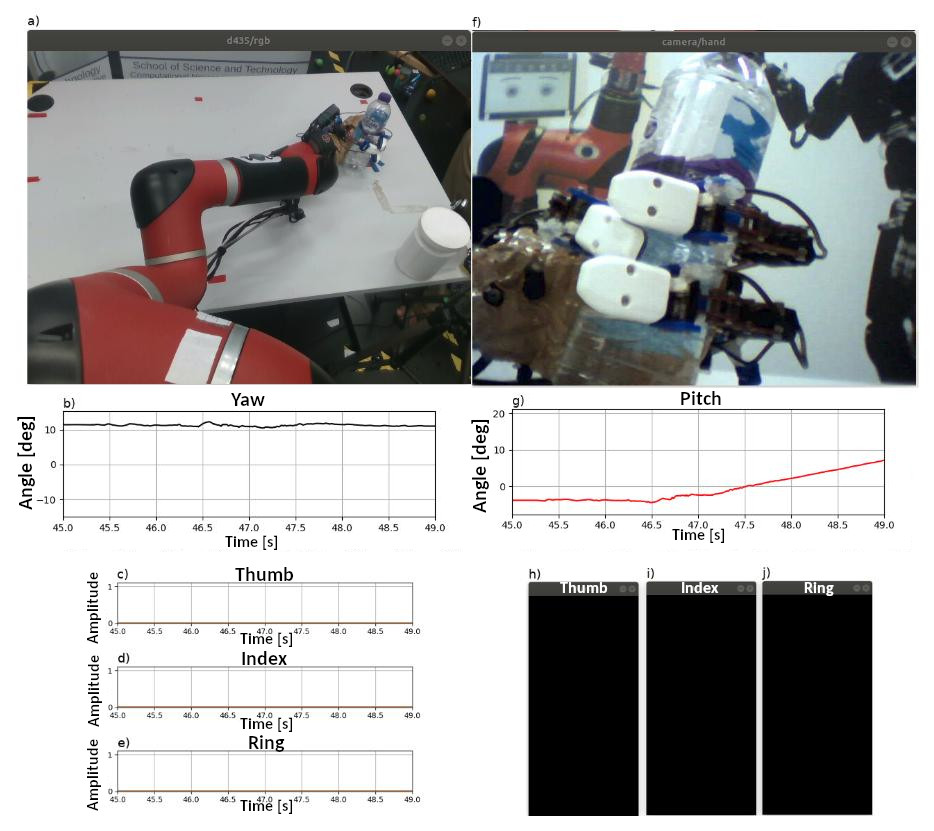}
	\caption{WTS-FT fingertips when grasping the bottle full of water (510g) during the grasping and lift phases. a) top view of the grasp at timestamp 47s, b) horizontal vibrations, c) plot of the Thumb fingertip's sensors, d) plot of the Index fingertip's sensors, e) plot of the Ring fingertip's sensors, f) front view at timestamp 47s, g) vertical vibrations, h) Thumb fingertip's sensors at timestamp 47s, i) Index fingertip's sensors at timestamp 47s, and j) Ring fingertip's sensors at timestamp 47s.}  \label{fig:wts_bottle}
	\end{center}
\end{figure}

Figure~\ref{fig:wts_bottle} shows the recordings when grasping the bottle full of water using the WTS-FT fingertips 
during the period of [45,46]s; the object was grasped because the water bottle was pushed against the palm of the robotic hand and the minimum joint thresholds reached (\textit{i.e.} the joints reached the minimum joint positions). From the images a) and f) it is possible to infer that the sensors are touching the water bottle but from Figure 10c,10d,10e it is clear the sensors are not recording any reading. This is a consequence of the low sensitivity of the WTS-FT fingertips. None of the fingertips' sensors (c, d and e) detected the bottle.

\begin{figure*}
	\begin{center}
	\includegraphics[width=1.0\textwidth]{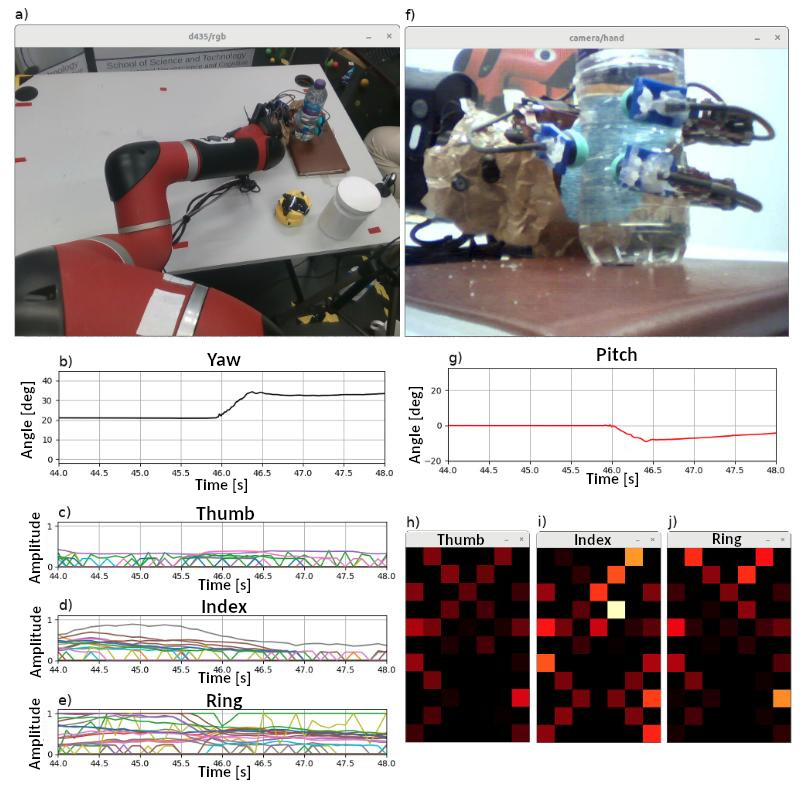}
	\caption{BioTac SP fingertips when grasping the bottle full of water (510g) during the grasping and lift phases. a) top view of the grasp at timestamp 46s, b) horizontal vibrations, c) plot of the Thumb fingertip's sensors, d) plot of the Index fingertip's sensors, e) plot of the Ring fingertip's sensors, f) front view at timestamp 47s, g) vertical vibrations, h) Thumb fingertip's sensors at timestamp 46s, i) Index fingertip's sensors at timestamp 46s, and j) Ring fingertip's sensors at timestamp 46s.}  \label{fig:biotac_bottle}
	\end{center}
\end{figure*}

Figure~\ref{fig:biotac_bottle} shows the equivalent recordings when grasping the bottle full of water using the BioTac SP fingertips. 
Figure~\ref{fig:wts_bottle} shows the object grasping which occurred during the period of [44,48]s; at least five sensors of each fingertip detected contact at 46s.The lift phase started 46s. The spatial representation of the three fingers show that the majority of the sensors are detecting the object. 

\section{Results analysis} \label{chap5:analysis}
The tests shown in Table~\ref{tab:perturbation_tests} show that the BioTac SP fingertips performed better grasping the can, cube box and tea cup tests and had the same performance results in the plastic bottle and plastic cup test. We conclude that the BioTac fingertips allow for better grasping of the heavier objects. The WTS-FT fingertips have proven to be less sensitive than the Biotac SP fingertips (see Table \ref{tab:perturbation_tests}). The main issues are related to the following:
\begin{enumerate}
    \item Size of the sensors (37.5 x 23.3 x 18.9 mm) - the WTS-FT are twice the size of the Biotac sensors and this therefore makes the fingers very long, which increases the complexity in closing the hand.
    \item Weight (15g) - The AR 10 robotic hand has to be calibrated more often with the WTS-FT fingertips because the weight of the WTS-FT require that the linear actuators (joints) work under more stress and therefore it was required to calibrate the AR10 more often when using the WTS-FT (about ten times) than the BioTac SP sensors (only one calibration required).
    \item Shape of sensor (flat) - this is the most relevant difference between the fingertips. The WTS-FT fingertips have no sensitivity on the edges, which prevents detection of small objects that slip between the fingers (\textit{e.g.} soft ball or the black cylinder). The lack of edge detection of the WTS-FT also impacts on the stress applied to the linear actuators (joints) of the AR10 hand. During the experiments the authors had to replace 2 linear actuators that stopped working, because of excess force applied to close the fingers when objects were being grasped with the edges of the WTS-FT sensors.
    \item Surface material (rubber) - unlike the BioTac sensors, which use a skin like material (with a rugosity similar to human skin) with more adherence to the surface of the object, the WTS-FT have a rubber-like material which decreases the slip resistance.
\end{enumerate}
Crucially, the experiments show that the simple control algorithm based on a proportional controller is quite capable of ensuring a robust grasp even when the object being grasped is significantly perturbed by shaking. This is the case for all the objects assessed. We conclude that this is only possible due to the sophistication and placements of the fingertip sensors, particularly in the case of the BioTac SP devices.

\section{Discussion and Conclusion} \label{chap6:discussion}
This paper has considered the potential of modern tactile sensors in simplifying the requirements for adaptive grasp algorithms, compatible with different types of haptic technologies. A simple proportional controller based adaptive grasp algorithm was used while performing a comparative analysis between two state-of-the-art commercially available fingertips, namely, the BioTac SP and WTS-FT fingertips. A thorough analysis of sensor contacts' strength, while grasping various objects, is provided. The results of this study show that the basic adaptive grasp algorithm is only limited by the  quality of the sensory data collected by each fingertip type. The BioTac devices performed better than the WTS-FT in the majority of the tests. The geometry, skin-like material, variety of embedded sensor type, and data acquisition speed of the BioTac fingertips are all factors that contribute to such successful results. The WTS-FT fingertips were unable to pick up heavier objects, during a shaking test, and the heavy objects slipped more than when using the BioTac sensors. We consider that this arises, at least in part, because the control algorithm is starved of appropriate sensor feedback  in the case of the WTS-FT sensors. We also conclude that WTS-FT fingertips are not suitable for grasping soft objects and are less tolerant to vibrations. Conversely, it is important to note that the WTS-FT fingertips are considerably less expensive than the BioTac SP fingertips. 

The following ROS stacks were developed in this project:\begin{enumerate}
    \item Sawyer Grasping project. The source code is available in the authors restricted repository \cite{machado_2019_sawyer}.
    \item WTS-FT. The source code is publicly available in \cite{machado_2019_wts_ft}.
    \item BioTac SP. The source code is publicly available \cite{machado_2019_biotac}.
    \item Object detection. The source code is stored on a private repository \cite{machado_2019_object_detection}.
\end{enumerate}

\section{Future Work} \label{chap7:future}
Future work is targeted at more extensive comparisons of tactile sensing fingertips and exploiting the BioTac sensors for tasks requiring fine object robot manipulation, i.e. tasks that can be easily  performed by humans using the sophisticated tactile sensing capabilities of the human hand (e.g. tie different types of knots). Exploiting advanced touch capabilities to inform modern robot perception algorithms will also be an important future challenge.

\bibliographystyle{frontiersinSCNS_ENG_HUMS} 
\bibliography{references}

\begin{thebibliography}{35}
\providecommand{\natexlab}[1]{#1}
\expandafter\ifx\csname urlstyle\endcsname\relax
  \providecommand{\doi}[1]{doi:\discretionary{}{}{}#1}\else
  \providecommand{\doi}{doi:\discretionary{}{}{}\begingroup
  \urlstyle{rm}\Url}\fi
\providecommand{\selectlanguage}[1]{\relax}
\providecommand{\bibAnnoteFile}[1]{%
  \IfFileExists{#1}{\begin{quotation}\noindent\textsc{Key:} #1\\
  \textsc{Annotation:}\ \input{#1}\end{quotation}}{}}
\providecommand{\bibAnnote}[2]{%
  \begin{quotation}\noindent\textsc{Key:} #1\\
  \textsc{Annotation:}\ #2\end{quotation}}

\bibitem[{Active-8-Robots(2020. [Online])}]{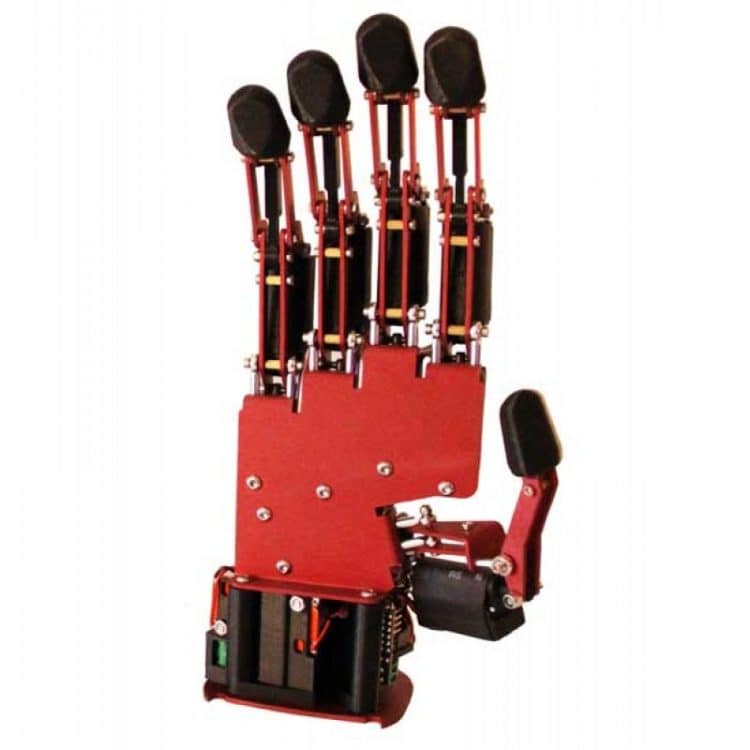}
Active-8-Robots (2020. [Online]).
\newblock Ar10 robotic hand.
\newblock Online
\bibAnnoteFile{ar10}

\bibitem[{Bhandari and Lee(2019)}]{Bhandari2019}
Bhandari, B. and Lee, M. (2019).
\newblock {Haptic identification of objects using tactile sensing and computer
  vision}.
\newblock \emph{Advances in Mechanical Engineering} 11, 168781401984046.
\newblock \doi{10.1177/1687814019840468}
\bibAnnoteFile{Bhandari2019}

\bibitem[{Chi et~al.(2018)Chi, Sun, Xue, Li, and Liu}]{Chi2018}
Chi, C., Sun, X., Xue, N., Li, T., and Liu, C. (2018).
\newblock {Recent Progress in Technologies for Tactile Sensors}.
\newblock \emph{Sensors} 18, 948.
\newblock \doi{10.3390/s18040948}
\bibAnnoteFile{Chi2018}

\bibitem[{Chorley et~al.(2009)Chorley, Melhuish, Pipe, and
  Rossiter}]{Chorley2009}
Chorley, C., Melhuish, C., Pipe, T., and Rossiter, J. (2009).
\newblock {Development of a tactile sensor based on biologically inspired edge
  encoding}.
\newblock \emph{2009 International Conference on Advanced Robotics, ICAR 2009}
  , 1--6
\bibAnnoteFile{Chorley2009}

\bibitem[{{Gomez Eguiluz} et~al.(2018){Gomez Eguiluz}, Rano, Coleman, and
  McGinnity}]{GomezEguiluz2018}
{Gomez Eguiluz}, A., Rano, I., Coleman, S., and McGinnity, T. (2018).
\newblock {Multimodal Material Identification through Recursive Tactile
  Sensing}.
\newblock \emph{IEEE Robotics and Autonomous Systems}
\bibAnnoteFile{GomezEguiluz2018}

\bibitem[{{G{\'{o}}mez Egu{\'{i}}luz} et~al.(2019){G{\'{o}}mez Egu{\'{i}}luz},
  Ra{\~{n}}{\'{o}}, Coleman, and McGinnity}]{GomezEguiluz2019}
{G{\'{o}}mez Egu{\'{i}}luz}, A., Ra{\~{n}}{\'{o}}, I., Coleman, S.~A., and
  McGinnity, T. (2019).
\newblock {Reliable robotic handovers through tactile sensing}.
\newblock \emph{Autonomous Robots} \doi{10.1007/s10514-018-09823-2}
\bibAnnoteFile{GomezEguiluz2019}

\bibitem[{Government(2018)}]{Government2018}
Government, U. (2018).
\newblock {Industrial Strategy}
\bibAnnoteFile{Government2018}

\bibitem[{Intel(2020. [Online])}]{Intel}
Intel (2020. [Online]).
\newblock Depth camera d435.
\newblock Online
\bibAnnoteFile{Intel}

\bibitem[{Kappassov et~al.(2015)Kappassov, Corrales, and
  Perdereau}]{Kappassov2015}
Kappassov, Z., Corrales, J.-A., and Perdereau, V. (2015).
\newblock Tactile sensing in dexterous robot hands — review.
\newblock \emph{Robotics and Autonomous Systems} 74, 195--220.
\newblock \doi{https://doi.org/10.1016/j.robot.2015.07.015}
\bibAnnoteFile{Kappassov2015}

\bibitem[{Ke et~al.(2019)Ke, Huang, Chen, Gao, Han, Wang et~al.}]{Ke2019}
Ke, A., Huang, J., Chen, L., Gao, Z., Han, J., Wang, C., et~al. (2019).
\newblock {Fingertip Tactile Sensor with Single Sensing Element Based on FSR
  and PVDF}.
\newblock \emph{IEEE Sensors Journal} , 1--1\doi{10.1109/JSEN.2019.2936304}
\bibAnnoteFile{Ke2019}

\bibitem[{Kerr et~al.(2013)Kerr, McGinnity, and Coleman}]{Kerr2013}
Kerr, E., McGinnity, T., and Coleman, S. (2013).
\newblock {Material classification based on thermal properties - A robot and
  human evaluation}.
\newblock In \emph{2013 IEEE International Conference on Robotics and
  Biomimetics, ROBIO 2013}.
\newblock \doi{10.1109/ROBIO.2013.6739602}
\bibAnnoteFile{Kerr2013}

\bibitem[{Kerr et~al.(2014)Kerr, McGinnity, and Coleman}]{Kerr2014}
Kerr, E., McGinnity, T., and Coleman, S. (2014).
\newblock {Material classification based on thermal and surface texture
  properties evaluated against human performance}.
\newblock In \emph{2014 13th International Conference on Control Automation
  Robotics and Vision, ICARCV 2014}.
\newblock \doi{10.1109/ICARCV.2014.7064346}
\bibAnnoteFile{Kerr2014}

\bibitem[{Kerr et~al.(2018)Kerr, McGinnity, and Coleman}]{Kerr2018}
Kerr, E., McGinnity, T., and Coleman, S. (2018).
\newblock Material recognition using tactile sensing.
\newblock \emph{Expert Systems with Applications}
  \doi{10.1016/j.eswa.2017.10.045}
\bibAnnoteFile{Kerr2018}

\bibitem[{Machado and McGinnity(2020{\natexlab{a}})}]{machado_2019_biotac_cube}
Machado and McGinnity (2020{\natexlab{a}}).
\newblock Enhanced grasp biotac sp cube trial 1.
\newblock \doi{10.5281/zenodo.3547726}
\bibAnnoteFile{machado_2019_biotac_cube}

\bibitem[{Machado and McGinnity(2020{\natexlab{b}})}]{machado_2019_wts-ft_cube}
Machado and McGinnity (2020{\natexlab{b}}).
\newblock Enhanced grasp wts-ft cube trial 1.
\newblock \doi{10.5281/zenodo.3547596}
\bibAnnoteFile{machado_2019_wts-ft_cube}

\bibitem[{Machado et~al.(2020{\natexlab{a}})Machado, Lama, and
  McGinnity}]{machado_2019_sawyer}
Machado, P., Lama, N., and McGinnity, T. (2020{\natexlab{a}}).
\newblock Sawyer grasping project.
\newblock https://doi.org/10.5281/zenodo.3529784.
\newblock \doi{10.5281/zenodo.3529784}
\bibAnnoteFile{machado_2019_sawyer}

\bibitem[{Machado et~al.(2020{\natexlab{b}})Machado, Lama, and
  McGinnity}]{machado_2019_wts_ft}
Machado, P., Lama, N., and McGinnity, T. (2020{\natexlab{b}}).
\newblock Wts-ft ros package.
\newblock https://doi.org/10.5281/zenodo.3529664.
\newblock \doi{10.5281/zenodo.3529664}
\bibAnnoteFile{machado_2019_wts_ft}

\bibitem[{Machado and McGinnity(2020{\natexlab{c}})}]{machado_2019_biotac}
Machado, P. and McGinnity, T. (2020{\natexlab{c}}).
\newblock Biotac sp ros stack.
\newblock https://doi.org/10.5281/zenodo.3529376.
\newblock \doi{10.5281/zenodo.3529376}.
\newblock N/A
\bibAnnoteFile{machado_2019_biotac}

\bibitem[{Machado and
  McGinnity(2020{\natexlab{d}})}]{machado_2019_object_detection}
Machado, P. and McGinnity, T. (2020{\natexlab{d}}).
\newblock Object detection.
\newblock https://doi.org/10.5281/zenodo.3529770.
\newblock \doi{10.5281/zenodo.3529770}
\bibAnnoteFile{machado_2019_object_detection}

\bibitem[{Nagabandi et~al.(2019)Nagabandi, Konoglie, Levine, and
  Kumar}]{Nagabandi2019}
Nagabandi, A., Konoglie, K., Levine, S., and Kumar, V. (2019).
\newblock {Deep Dynamics Models for Learning Dexterous Manipulation}.
\newblock In \emph{Robotics: Science and Systems 2018}. 1--12
\bibAnnoteFile{Nagabandi2019}

\bibitem[{OpenCV(2020. [Online])}]{OpenCV}
OpenCV (2020. [Online]).
\newblock Opencv.
\newblock Online
\bibAnnoteFile{OpenCV}

\bibitem[{Ottenhaus et~al.(2016)Ottenhaus, Miller, Schiebener, Vahrenkamp, and
  Asfour}]{Ottenhaus2016}
Ottenhaus, S., Miller, M., Schiebener, D., Vahrenkamp, N., and Asfour, T.
  (2016).
\newblock {Local implicit surface estimation for haptic exploration}.
\newblock In \emph{2016 IEEE-RAS 16th International Conference on Humanoid
  Robots (Humanoids)} (IEEE), 850--856.
\newblock \doi{10.1109/HUMANOIDS.2016.7803372}
\bibAnnoteFile{Ottenhaus2016}

\bibitem[{Ottenhaus et~al.(2018)Ottenhaus, Weiner, Kaul, Tulbure, and
  Asfour}]{Ottenhaus2018}
Ottenhaus, S., Weiner, P., Kaul, L., Tulbure, A., and Asfour, T. (2018).
\newblock {Exploration and Reconstruction of Unknown Objects using a Novel
  Normal and Contact Sensor}.
\newblock In \emph{2018 IEEE/RSJ International Conference on Intelligent Robots
  and Systems (IROS)} (IEEE), 1614--1620.
\newblock \doi{10.1109/IROS.2018.8594272}
\bibAnnoteFile{Ottenhaus2018}

\bibitem[{{Raspberry Pi Foundation}(2020)}]{RaspberryPiFoundation}
{Raspberry Pi Foundation} (2020).
\newblock {Buy a Raspberry Pi 3 Model B – Raspberry Pi}.
\newblock https://www.raspberrypi.org/products/raspberry-pi-3-model-b/
\bibAnnoteFile{RaspberryPiFoundation}

\bibitem[{Realsense(2020. [Online])}]{Realsense}
Realsense (2020. [Online]).
\newblock Intelrealsense/librealsense.
\newblock Online
\bibAnnoteFile{Realsense}

\bibitem[{Rethink-Robotics(2020. [Online])}]{Sawyer}
Rethink-Robotics (2020. [Online]).
\newblock Sawyer robot.
\newblock Online
\bibAnnoteFile{Sawyer}

\bibitem[{Robotics(2020. [Online])}]{RethinkRobotics}
Robotics, R. (2020. [Online]).
\newblock Workstation setup - intera sdk.
\newblock Online
\bibAnnoteFile{RethinkRobotics}

\bibitem[{ROS(2020. [Online])}]{ROS}
ROS (2020. [Online]).
\newblock Ros kinetic.
\newblock Online
\bibAnnoteFile{ROS}

\bibitem[{Rouhafzay and Cretu(2019)}]{Rouhafzay2019}
Rouhafzay, G. and Cretu, A.-M. (2019).
\newblock {A Visuo-Haptic Framework for Object Recognition Inspired by Human
  Tactile Perception}.
\newblock \emph{Proceedings} 4, 47.
\newblock \doi{10.3390/ecsa-5-05754}
\bibAnnoteFile{Rouhafzay2019}

\bibitem[{Saudabayev et~al.(2018)Saudabayev, Rysbek, Khassenova, and
  Varol}]{Saudabayev2018}
Saudabayev, A., Rysbek, Z., Khassenova, R., and Varol, H.~A. (2018).
\newblock {Human grasping database for activities of daily living with depth,
  color and kinematic data streams}.
\newblock \emph{Scientific Data} 5, 180101.
\newblock \doi{10.1038/sdata.2018.101}
\bibAnnoteFile{Saudabayev2018}

\bibitem[{Syntouch(2020. [Online])}]{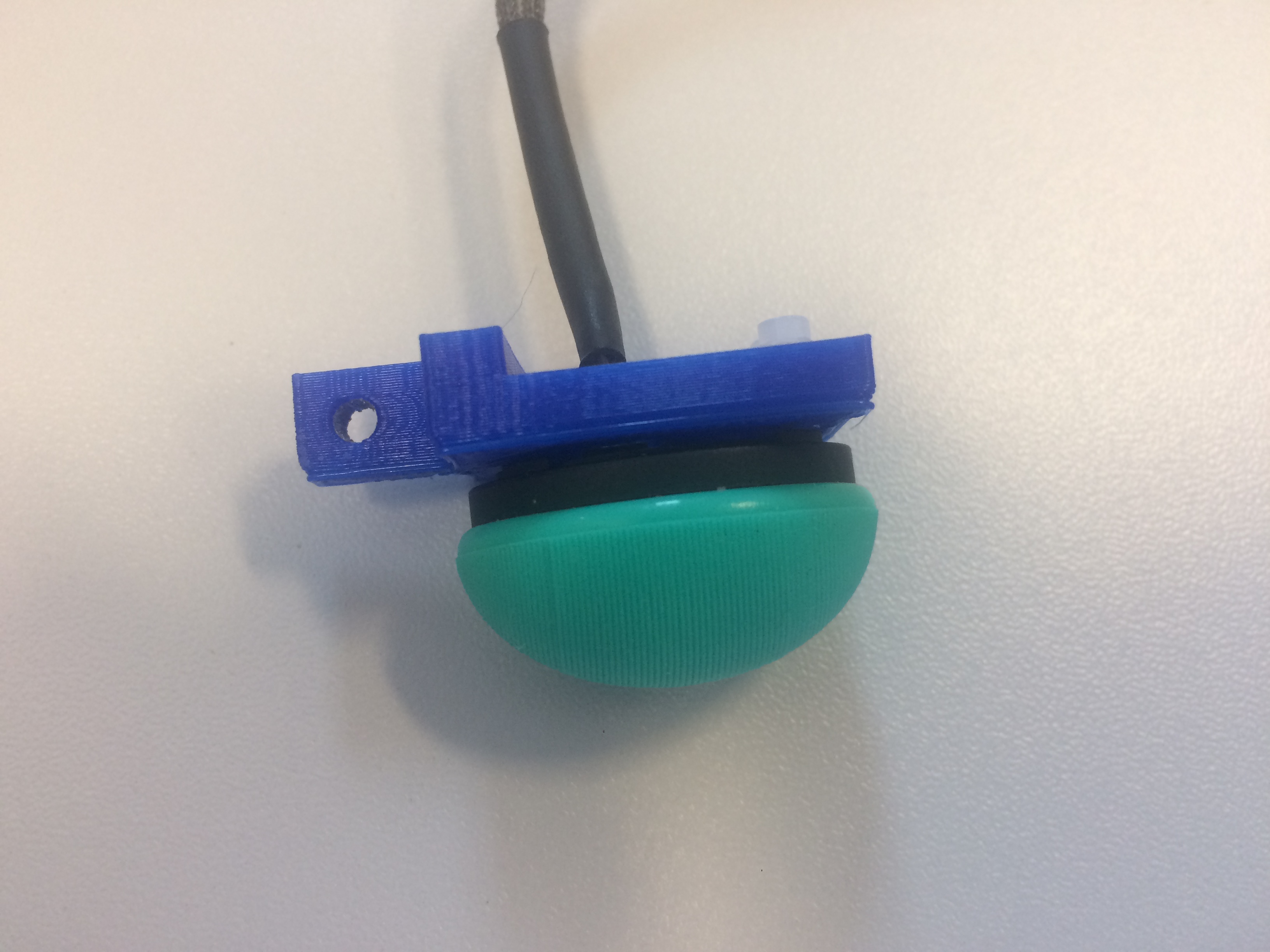}
Syntouch (2020. [Online]).
\newblock Biotac sp fingertip.
\newblock Online
\bibAnnoteFile{biotac}

\bibitem[{Takktile(2020. [Online])}]{Takktile}
Takktile (2020. [Online]).
\newblock Takkstrip basics - takktile.
\newblock Online
\bibAnnoteFile{Takktile}

\bibitem[{Ubuntu(2020. [Online])}]{Ubuntu}
Ubuntu (2020. [Online]).
\newblock Ubuntu 16.04.6 lts (xenial xerus).
\newblock Online
\bibAnnoteFile{Ubuntu}

\bibitem[{Weiss-Robotics(2020. [Online])}]{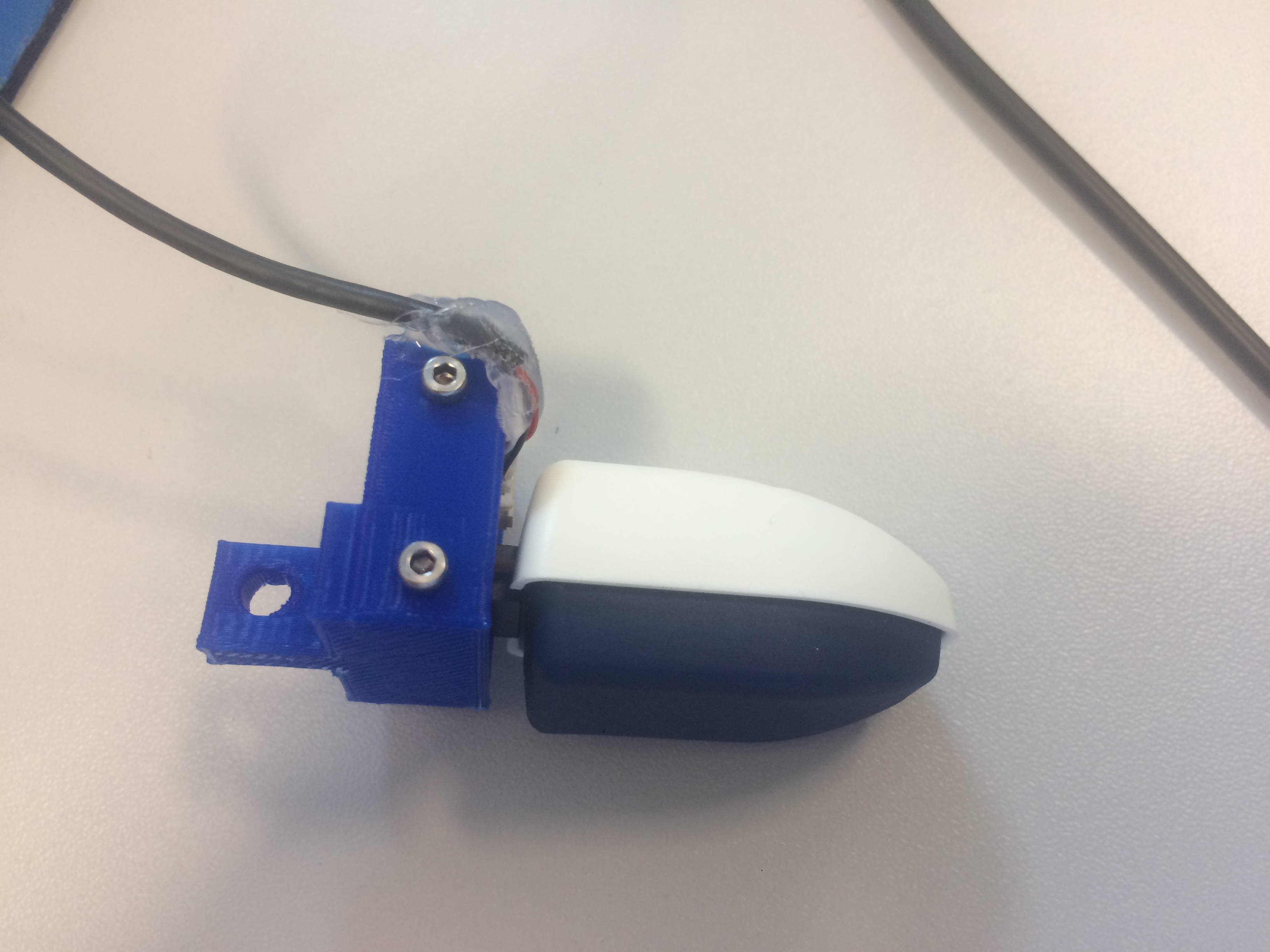}
Weiss-Robotics (2020. [Online]).
\newblock Wts-ft fingertip.
\newblock Online
\bibAnnoteFile{wts-ft}

\bibitem[{Yeon et~al.(2017)Yeon, Kim, Ryu, Park, Chung, and Kim}]{Yeon2017}
Yeon, J., Kim, J., Ryu, J., Park, J.-Y., Chung, S.-C., and Kim, S.-P. (2017).
\newblock {Human Brain Activity Related to the Tactile Perception of
  Stickiness}.
\newblock \emph{Frontiers in Human Neuroscience} 11, 8.
\newblock \doi{10.3389/fnhum.2017.00008}
\bibAnnoteFile{Yeon2017}

\end{thebibliography}

\end{document}